# Wall Inspector: Quadrotor Control in Wall-proximity Through Model Compensation

Peiwen Yang*, Weisong Wen*, Runqiu Yang*, Yingming Chen*, and Cheuk Chi Tsang

*These authors are with the Department of Aeronautical and Aviation Engineering, Hong Kong Polytechnic University, Hong Kong; Cheuk Chi Tsang is with University of California Berkeley (e-mail: peiwen1.yang@connect.polyu.hk; welson.wen@polyu.edu.hk; runqiu.yang@polyu.edu.hk; yingming.chen@polyu.edu.hk; mclaren_tsang@berkeley.edu) (Corresponding author: Weisong, Wen.)

*Abstract*— The safe operation of quadrotors in near-wall urban or indoor environments (e.g., inspection and search-and-rescue missions) is challenged by unmodeled aerodynamic effects arising from wall-proximity. It generates complex vortices that induce destabilizing suction forces, potentially leading to hazardous vibrations or collisions. This paper presents a comprehensive solution featuring (1) a physics-based suction force model that explicitly characterizes the dependency on both rotor speed and wall distance, and (2) a suction-compensated model predictive control (SC-MPC) framework designed to ensure accurate and stable trajectory tracking during wall-proximity operations. The proposed SC-MPC framework incorporates an enhanced dynamics model that accounts for suction force effects, formulated as a factor graph optimization problem integrating system dynamics constraints, trajectory tracking objectives, control input smoothness requirements, and actuator physical limitations. The suction force model parameters are systematically identified through extensive experimental measurements across varying operational conditions. Experimental validation demonstrates SC-MPC's superior performance, achieving 2.1 cm root mean squared error (RMSE) in X-axis and 2.0 cm RMSE in Y-axis position control - representing 74% and 79% improvements over cascaded proportional-integral-derivative (PID) control, and 60% and 53% improvements over standard MPC respectively. The corresponding mean absolute error (MAE) metrics (1.2 cm X-axis, 1.4 cm Y-axis) similarly outperform both baselines. The evaluation platform employs a ducted quadrotor design that provides collision protection while maintaining aerodynamic efficiency. To facilitate reproducibility and community adoption, we have open-sourced our complete implementation, including control system codebase, experimental datasets, and demonstration videos, available at https://anonymous.4open.science/r/SC-MPC-6A61.

| NOMENCLATURE | | |
|---|---|---|
| $\{P_k\}$ | $k$-th plane frame | |
| $\{w\}$ | world frame | |
| $\{b\}$ | body frame | |
| $\mathbf{p}$ | position rotation | m |
| $\mathbf{R}$ | rotation matrix | |
| $\mathbf{v}$ | velocity | m/s |
| $\boldsymbol{\omega}_b$ | angular velocity | rad/s |
| $\mathbf{F}_s$ | suction force in the world frame | N |
| $k_s$ | Suction force scale factor | N/m |
| $d_{thr}$ | threshold distance | m |
| $d$ | distance from the center of the rotor to the wall | m |
| $T_b$ | thrust in the body frame | N |
| $\mathbf{F}_{drag}^b$ | drag force in the body frame | N |
| $\mathbf{I}_b$ | moment of inertia | kg×m² |
| $\mathbf{M}_b$ | aerodynamic torque | N×m |
| $\mathbf{u}$ | control input | |
| $\mathbf{x}^r, \mathbf{u}^r$ | reference state and reference input | |
| $\mathbf{P}^D$ | covariance matrix of dynamic residuals | |
| $\mathbf{D}$ | aerodynamic drag coefficient matrix | N×s/m |
| g | gravitational acceleration | m/s² |
| $m_b$ | mass | kg |
| $\mathbf{p}_{r_j}^b$ | $j$-th rotor's position in the body frame | m |
| $\mathbf{Q}, \mathbf{Q}_N, \mathbf{P}_k^D, \mathbf{G}$, and $\mathbf{Q}_{lim}$ | weighting matrices in MPC | |
| $\mathbf{I}_3$ | 3-dimensional identify matrix | |

## 1. INTRODUCTION

Multi-rotor aerial vehicles (MAVs), particularly quadrotors, have increasingly been deployed in confined urban or indoor environments due to their maneuverability and versatility. These capabilities enable applications such as firefighting, search-and-rescue, external window cleaning, and tunnel or wall inspection [1-3], which often necessitate operations near vertical surfaces. However, in these constrained environments, aerodynamic disturbances significantly pose a significant challenge to MAV stability because of the wall-proximity effect. This effect induces complex flow interactions and unsteady aerodynamic forces. To mitigate these issues, this study introduces the 'Wall Inspector', a specialized system to actively estimate and counteract the complex aerodynamic disturbances induced by wall proximity, thereby enabling robust flight control in such environments.

A significant aerodynamic disturbance acting on the quadrotors operating near walls has been identified [4, 5]. Control experiments further indicate that both cascaded proportion-integral- derivative (PID) and model predictive control (MPC) [6, 7] struggle to maintain stability and prevent collision, as these disturbances cause the quadrotor to drift towards the wall. This resulting instability leads to repeated oscillations due to the wall-proximity effect. As shown in Fig. 1-a, airflow direction was visualized using smog [8]. Additionally, force measurement experiments conducted on a test bench, as depicted in Fig. 1-b, quantified the magnitude of the suction force and established its relationship with parameters such as wall distance and rotor speed.

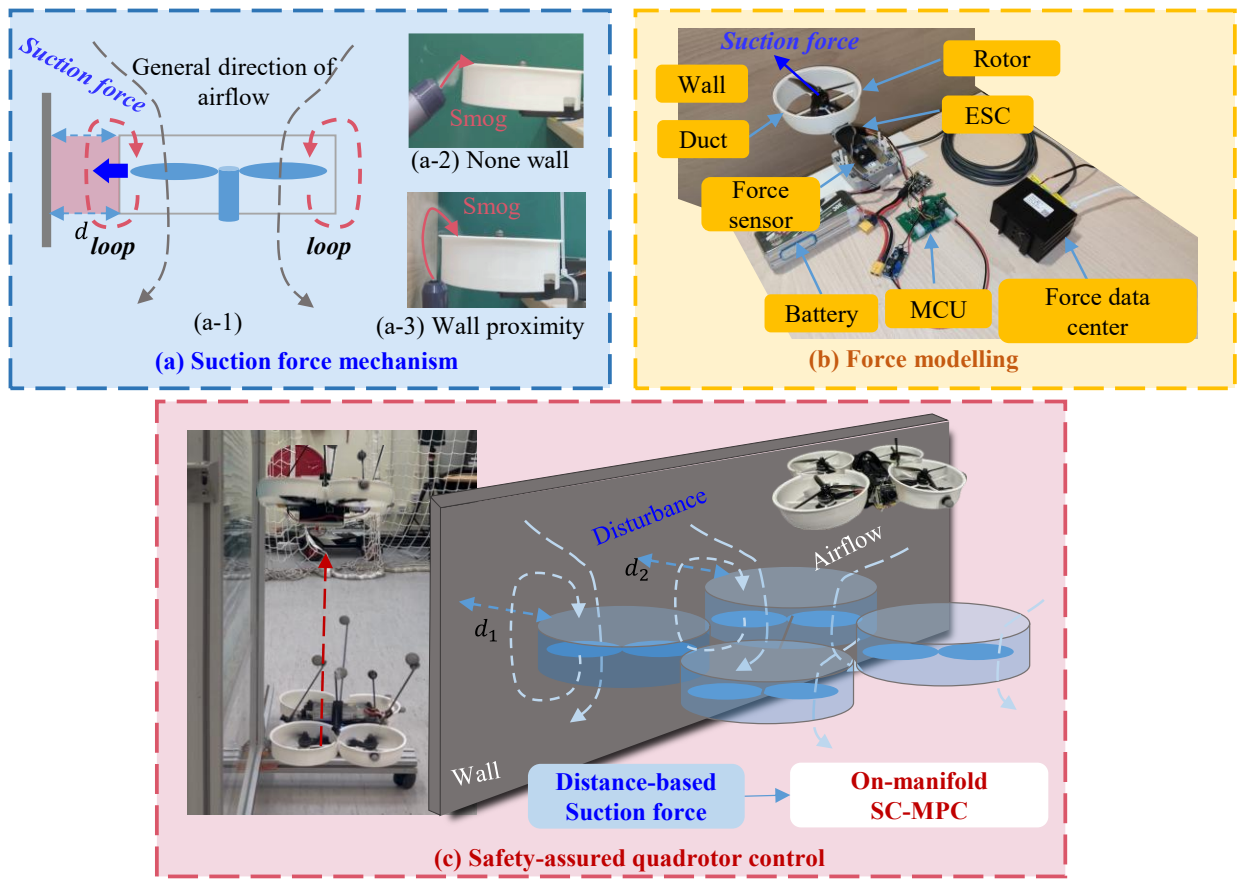

Fig. 1. Proposed accurate quadrotor control in wall-proximity with quantifiable aerodynamic effect investigation. (a) Investigation of the suction force mechanism: (a-1) A schematic drawing depicts the airflow around the rotor duct when a ducted rotor operates near a wall. (a-2) and (a-3) display the smog flow paths in free space and near a wall, respectively. (b) Force measurement experiments are conducted on a test bench. (c) The proposed control approach achieves accurate trajectory tracking and collision avoidance with walls. Video link: https://youtu.be/wjbxtRiHkPk.

Suction force modelling is vital to maintain trajectory accuracy by compensating for dynamics model errors in quadrotors near walls. As illustrated in Fig. 1-c, to fill this gap, this paper utilizes a more comprehensive dynamics model, that leverages distance measurements between the wall and the ducts to enhance the control methodology, thereby ensuring safer trajectory adherence in proximity to walls. In addition, offline parameters derived from test bench experiments exhibit variability during flight. To address this, parameter identification from flight data is implemented using the Levenberg-Marquardt algorithm [9, 10] to refine the suction force model parameters.

*Contributions*: This paper elucidates the wall-proximity effect on rotors through airflow visualization and empirical tabletop experiments, culminating in a suction force model. Furthermore, the suction-compensated dynamics model is integrated into an on-manifold model predictive control (MPC) framework. During dynamic flight near walls, a more accurate model is derived from flight data, accounting for differences in aerodynamic configurations between practical flight and tabletop experiments. The primary contributions are as follows:

(1) **A physics-based suction force model for wall-proximity aerodynamics:** We present a detailed exposition and the suction force model explicitly characterizing the dependence on both rotor speed and wall distance for ducted quadrotor rotors. This model is validated through systematic smog flow visualization and force measurement experiments. The exposition proposed through Bernoulli's principle is different from existing research [4, 5].

(2) **An aerodynamic parameter identification method for flight data**: We propose an aerodynamics parameter identification technique for refining the dynamics model. The model parameter is optimized using the Levenberg-Marquardt algorithm. This method explicitly addresses the discrepancy between static bench-test parameters and dynamic flight conditions by minimizing dynamics model residuals, enabling accurate model adaptation to real-world aerodynamic configurations encountered during operation.

(3) **A real-time suction-compensated on-manifold MPC (SC-MPC) framework**: We propose a real-time capable (100 Hz) model predictive controller integrating the identified suction model into an on-manifold optimization formulation. By formulating the optimization problem, the nonlinear iterative optimization is implemented that inherently reduces linearization errors. Explicitly compensating for wall-proximity disturbances within the dynamics factor residuals enables robust trajectory tracking and collision avoidance.

(4) **Comprehensive experimental validation with a ducted quadrotor platform**: Except for simulations, we demonstrate the superior performance of SC-MPC through near-wall trajectory tracking experiments using a custom ducted quadrotor. Results show 74% and 79% RMSE reduction in X/Y-axis position control versus cascaded PID, and 60% and 53% versus standard MPC, while achieving collision-free operation where baselines fail. The complete system implementation is open-sourced for reproducibility.

## 2. RELATED WORK

*2.1 Dynamics Model and its Identification*

Numerous studies have explored the ideal dynamics model of quadrotor [11-14], all of which depend on idealized assumptions. For example, the Kalman filter has been employed to identify the dynamics model online for visual-inertial MAVs [14, 15] as well as small fixed-wing drones [16]. In high-speed scenarios, traditional models prove inadequate, prompting the use of deep learning to capture dynamics model residuals [17, 18]. The aerodynamic performance of a ducted-rotor system is influenced by the tip-gap distance, as demonstrated through thrust measurements and static wall-pressure analysis [19]. The study [20] investigated shaftless ducted rotor (SDR). However, all these methods have been validated primarily on static scenarios. For instance, surrounding walls seriously influence quadrotor dynamics. Thus, some scholars have addressed dynamics models in non-ideal environments, with [5] attributing the wall-proximity effect to rotor deflection caused thrust towards the wall. Additionally, [8] investigated the dynamic interactions of MAVs approaching each other's base, visualizing airflow characteristics with smog. Learning-based methods have also enhanced quadrotor control stability and accuracy [17, 18], though they rely heavily on data and lack interpretability.

*2.2 Quadrotor Control*

MPC is a predominant approach for real-time optimization-based planning and control [21-26]. It's renowned for its predictive ability and proficiency to handle constraints. To cope with the computation complexity of MPC, many studies have proposed accelerated methods to ensure the real-time performance of MPC [24, 27].

FGO is frequently utilized for positioning problems, with further details available in [28-30]. FGO optimizes variable increments depending on the last linearized point, often derived from trajectory planning. Optimal control and estimation are dual in the *linear-quadratic-Gaussian* (LQG) setting, as Kalman discovered, then Todorov [31] generalized this result to non-linear stochastic systems, discrete stochastic systems, and deterministic systems. For instance, maximum *a posteriori* (MAP) estimation and optimal control problem are dual in the non-linear stochastic systems [31]. Consequently, FGO has been effectively applied to MPC problem [32-36] by framing MPC as a MAP estimation task. Yang [33] addressed optimal control problems using constrained factor graphs and optimized the factor graphs to obtain the optimal trajectory and the feedback control policies using the variable elimination algorithm. Lu [23] formulated the MPC based on a canonical representation of on-manifold systems, which is linearized at each point along the incremental trajectory without singularities.

Reinforcement learning has also been applied to control problems for close proximity quadcopter flight [37]. However, RL suffers from high sample inefficiency (requiring extensive, costly real-world interactions), safety risks due to exploratory actions (e.g., drone crashes), and lack of theoretical guarantees (black-box policies with unverified stability).

## 3. MODELLING AND METHODOLOGY

*3.1 Operations*

The existence of homeomorphisms around any point in a manifold $\mathcal{M}$ enables the definition of two operators, $\boxplus$ and $\boxminus$, which provide a local, vectorized representation of the manifold's globally complex geometry. These operators are defined as follows [38]:

Addition operator: The $\boxplus$ operator maps a state on the manifold and a perturbation in the local Euclidean space to a new state on the manifold:

$$\boxplus: \mathcal{M} \times \mathbb{R}^n \to \mathcal{M}, \mathbf{x} \boxplus \mathbf{u} = \phi_\mathbf{x}^{-1}(\mathbf{u}), \forall \mathbf{u} \in B_\epsilon(\mathbf{x}) \tag{1}$$

where $\phi_\mathbf{x}: U_\mathbf{x} \to \mathbb{R}^n$ is a homeomorphism from a neighborhood $U_\mathbf{x} \subset \mathcal{M}$ around $\mathbf{x} \in \mathcal{M}$ to an open subset of $\mathbb{R}^n$, and $B_\epsilon(\mathbf{x}) \subset \mathbb{R}^n$ is the region where the homeomorphism is valid.

Subtraction operator: The $\boxminus$ operator computes the perturbation that, when added to a state, yields another state on the manifold:

$$\boxminus: \mathcal{M} \times \mathcal{M} \to \mathbb{R}^n, \mathbf{y} \boxminus \mathbf{x} = \phi_\mathbf{x}(\mathbf{y}), \forall \mathbf{y} \in U_\mathbf{x} \tag{2}$$

Physically, $\mathbf{y} = \mathbf{x} \boxplus \mathbf{u}$ represents adding a small perturbation $\mathbf{u} \in \mathbb{R}^n$ to the state $\mathbf{x} \in \mathcal{M}$ moving along the manifold to a new state $\mathbf{y}$. Conversely, $\mathbf{y} \boxminus \mathbf{x}$ determines the perturbation $\mathbf{u}$ that, when added to $\mathbf{x}$ via $\boxplus$, yields $\mathbf{y}$. These operators create a local Euclidean view of the manifold, facilitating control design and analysis.

*3.2 Dynamics Model*

The nominal dynamics of the quadrotor [13] are described by:

$$\begin{aligned} \dot{\mathbf{p}}_b^w &= \mathbf{v}_b^w \\ \dot{\mathbf{R}}_b^w &= \mathbf{R}_b^w \boldsymbol{\omega}_b^\times \\ \dot{\mathbf{v}}_b^w &= -\mathbf{e}_3 g + \frac{\mathbf{R}_b^w (\mathbf{e}_3 T_b + \mathbf{F}_{drag}^b)}{m_b} \\ \dot{\boldsymbol{\omega}}_b &= \mathbf{I}_b^{-1}(-\boldsymbol{\omega}_b^\times \mathbf{I}_b \boldsymbol{\omega}_b + \mathbf{M}_b) \end{aligned} \tag{3}$$

where $\mathbf{p}_b^w, \mathbf{R}_b^w, \mathbf{v}_b^w, \boldsymbol{\omega}_b$ represent the position, rotation, velocity, and angular velocity of the quadrotor, respectively. Here, $\mathbf{e}_3 = [0,0,1]^T$ is a unit vector, $m_b$ is the mass, g is the gravitational acceleration, $T_b$ is the thrust, $\mathbf{F}_{drag}^b$ denotes the aerodynamic drag, $\mathbf{I}_b$ is the inertia tensor, and $\mathbf{M}_b$ is the aerodynamic torque. The aerodynamic drag model is as follows:

$$\mathbf{F}_{drag}^b = \mathbf{D}\mathbf{R}_w^b \mathbf{v}_b^w \tag{4}$$

where $\mathbf{D} = \text{diag}([d_1, d_2, d_3]^T)$ represents the drag matrix.

The model (3) works well for non-interaction environments. However, during wall-proximity flight, the suction force significantly affects control stability and accuracy. When the quadrotor operates in close proximity to the wall, the acceleration model is modified to incorporate suction effects, resulting in the following expression:

$$\dot{\mathbf{v}}_b^w = -\mathbf{e}_3 g + \frac{\mathbf{R}_b^w(\mathbf{e}_3 T_b + \mathbf{F}_{drag}^b) + \mathbf{F}_s}{m_b} \tag{5}$$

where $\mathbf{F}_s$ represents the suction force due to the wall-proximity effect. As depicted in Fig. 2, a quadrotor operates in proximity to a planar surface $i$, where rotor 2 and rotor 3 are subject to a specific suction force. The underlying mechanism, as illustrated in Fig. 1-a, involves the formation of a closed airflow loop between the wall and the duct rotor. The high-velocity airflow in this region generates negative pressure, contributing to the suction effect.

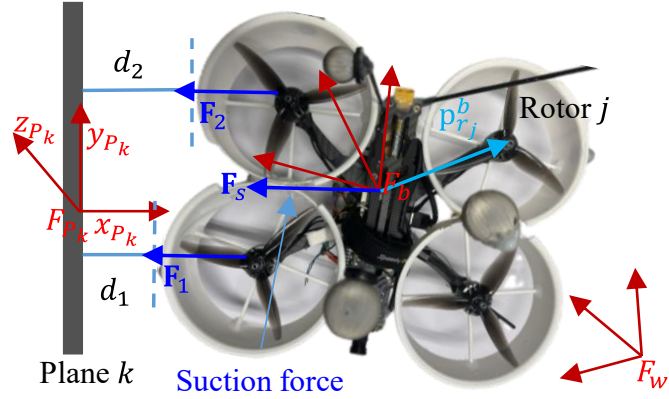

Fig. 2. Force modelling in wall-proximity. $F_{P_k}$, $F_b$, and $F_w$ represent the $k$-th plane frame, body frame, and world frame, respectively. The reference frame is defined such that the x-axis is normal to the wall and directed outward, the z-axis is antiparallel to gravity, and the y-axis follows the right-hand rule.

Force measurement experiments, as depicted in Fig. 1-b, are conducted to quantify the model. All experiments illustrating this theory will be detailed in Sec. IV through the force measurement experiments. The suction force $F_s$ is proportional to the wall distance, expressed as $F_s \propto (d_{thr} - d)$, and diminishes to zero when $d$ exceeds the threshold $d_{thr}$ [4, 5]. The suction force $F_s$ is mathematically defined as:

$$F_s = \begin{cases} k_s(d_{thr} - d) & d < d_{thr} \\ 0 & d \geq d_{thr} \end{cases} \quad (6)$$

where $k_s > 0$ and $d_{thr} > 0$ are model parameters.

The suction force in the current model is derived from a pressure differential induced by flow acceleration near the wall, consistent with Bernoulli's principle [39]. The gap flow velocity model is derived as follows: According to the suction model (6) and Bernoulli's principle [39], for a fluid of density $\rho$, when the local flow velocity $v(d)$ increases as the gap distance $d$ decreases, the resulting pressure drops,

$$\Delta P(d) = \frac{1}{2}\rho\left(v_{ref}^2 - v(d)^2\right) = \frac{1}{2}\kappa(d_{thr} - d) \quad (7)$$

which generates an attractive force proportional to $d_{thr} - d$ when $d < d_{thr}$, where $\kappa$ is a constant value. Here, $v_{ref}$ is the freestream velocity, and $v(d)$ is approximated as,

$$v(d) = v_{ref}\sqrt{1 + \frac{\kappa(d_{thr} - d)}{v_{ref}^2}} \approx v_{ref}\left(1 + \frac{\kappa(d_{thr} - d)}{2v_{ref}^2}\right) \tag{8}$$

Beyond $d_{thr}$, the pressure equilibrates ($\Delta P = 0$), terminating the force and $v(d) = v_{ref}$. The aerodynamic suction force is first modeled for a single ducted rotor. Since the quadrotor is equipped with four such ducted rotors, each rotor's suction force must be defined in the appropriate frames. Let $\mathbf{p}_{r_j}^b$ denote the $j$-th rotor's position in the body frame. Then, the position of the $j$-th rotor in the $k$-th plane frame is expressed as follows:

$$\mathbf{p}_{r_j}^{P_k} = \mathbf{R}_w^{P_k}\left(\mathbf{R}_b^w \mathbf{p}_{r_j}^b + \mathbf{p}_b^w\right) + \mathbf{p}_w^{P_k} \tag{9}$$

Therefore, the suction force $\mathbf{F}_s$ model of the quadrotor due to the wall-proximity effect is as follows:

$$\mathbf{F}_s = \mathbf{R}_{P_k}^w \sum_{j=0}^{3} \mathbf{e}_1 k_s d_j,$$

$$d_j = \begin{cases} \left|\mathbf{e}_1^T \mathbf{p}_{r_j}^{P_k}\right| - d_{thr} & d_{thr} - \mathbf{e}_1^T \mathbf{p}_{r_j}^{P_k} \geq 0 \\ 0 & d_{thr} - \mathbf{e}_1^T \mathbf{p}_{r_j}^{P_k} < 0 \end{cases} \tag{10}$$

where $\mathbf{e}_1 = [1, 0, 0]^T$.

*3.3 Suction-compensated Model Predictive Control*

The control framework contains two layers. The upper-level layer is an MPC, which takes thrust and angular velocity as the control input $\mathbf{u} = [T_b, \boldsymbol{\omega}_b^T]^T$. Its state $\mathbf{x} = \left[\left(\mathbf{p}_b^w\right)^T, \left(\boldsymbol{\theta}_b^w\right)^T, \left(\mathbf{v}_b^w\right)^T\right]^T$, where $\boldsymbol{\theta}_b^w$ is a rotation vector corresponding to a rotation matrix $\mathbf{R}_b^w$, i.e., $\exp\left\{\left(\boldsymbol{\theta}_b^w\right)^\wedge\right\} = \mathbf{R}_b^w$. The low-level layer achieves thrust and angular velocity tracking from the output of MPC.

The objective of MPC is to minimize the quadratic penalty associated with the deviation between the predicted states and their respective reference values over a defined future time horizon $N$ [6]. The traditional MPC problem can be formulated as follows:

$$\begin{aligned}
\min_{\mathbf{u}_{0:N-1}} &\sum_{k=0}^{N-1}\left(\|\mathbf{x}_k - \mathbf{x}_k^r\|_\mathbf{Q}^2 + \|\mathbf{u}_k - \mathbf{u}_k^r\|_\mathbf{G}^2\right) + \|\mathbf{x}_N - \mathbf{x}_N^r\|_\mathbf{P}^2 \\
\text{s.t.} \quad & \mathbf{x}_{k+1} = \mathbf{F}(\mathbf{x}_k, \mathbf{u}_k), \ 0 \leq k \leq N-1 \\
& \mathbf{u}_{min} \leq \mathbf{u}_k \leq \mathbf{u}_{max}, \\
& \mathbf{x}_k \in \chi, \mathbf{u}_k \in U, \ 0 \leq k \leq N-1 \\
& \mathbf{x}_N \in \chi_f \\
& \mathbf{x}_0 = \mathbf{x}_{init}
\end{aligned} \tag{11}$$

where $\chi \subseteq \mathbb{R}^n$ and $\mathcal{U} \subseteq \mathbb{R}^m$ are polyhedral sets, $\chi_f \subseteq \mathbb{R}^n$ is a terminal polyhedral region, $\mathbf{Q} \geq 0$, $\mathbf{G} > 0$, and $\mathbf{P} \geq 0$ are the state, input, and final state weighting matrices, respectively, the $\mathbf{u}_k$ and $\mathbf{u}_k^r$ ( $0 \leq k \leq N-1$) are the control input and the reference input, respectively, $\mathbf{x}_k^r (1 \leq k \leq N-1)$ and $\mathbf{x}_N^r$ are reference states, and $\mathbf{x}_{init}$ is the initial state from the state estimator.

To obtain the solution of MPC using FGO [32, 40], the problem (11) can be reformulated as follows:

$$\min_{u_{0:N-1}} f_c = \begin{Bmatrix} \sum_{k=1}^{N} \left(\mathbf{x}_k \boxminus \mathbf{z}_k^r\right)_{\mathbf{Q}}^2 + \left(\mathbf{x}_k \boxminus \mathbf{z}_N^r\right)_{\mathbf{Q}_N}^2 \\ +\sum_{k=1}^{N-1} \left(\mathbf{x}_{k+1} \boxminus \mathbf{F}(\mathbf{x}_k, \mathbf{u}_k)\right)_{\mathbf{P}_k^D}^2 \\ +\sum_{k=0}^{N-2} \left(\mathbf{u}_k \boxminus \mathbf{u}_{k+1}\right)_{\mathbf{G}}^2 + \sum_{k=0}^{N-1} \mathbf{h}\left(\mathbf{u}_k\right)_{\mathbf{Q}_{lim}}^2 \end{Bmatrix} \quad (12)$$

$$\text{s.t.} \quad \mathbf{x}_0 = \mathbf{x}_{init}$$

where $\mathbf{Q}$, $\mathbf{Q}_N$, $\mathbf{P}_k^D$, $\mathbf{G}$, and $\mathbf{Q}_{lim}$ represent different weighting matrices.

We define all these penalty terms in a cost function, such as trajectory tracking penalty term and smoothness penalty term. The control input rate $\mathbf{r}^U(\mathbf{u}_k, \mathbf{u}_{k+1}) = \mathbf{u}_k \boxminus \mathbf{u}_{k+1}$ is restricted to prevent changes in the control input from exceeding the capacity of the actuator. Moreover, the input adheres to the bound limit function $\mathbf{h}(\mathbf{u}_k)$. Additionally, the control part involves the reference trajectory factor residuals $\mathbf{r}^{ref}(\mathbf{x}, \mathbf{x}^{ref}) = \mathbf{x} \boxminus \mathbf{x}^{ref}$, and dynamic factor residuals $\mathbf{r}^D(\mathbf{x}_{k+1}, \mathbf{x}_k, \mathbf{u}_k) = \mathbf{x}_{k+1} \boxminus \mathbf{F}(\mathbf{x}_k, \mathbf{u}_k)$. The factor graph of MPC can be seen in Fig. 3. The factors will be discussed in detail in the following.

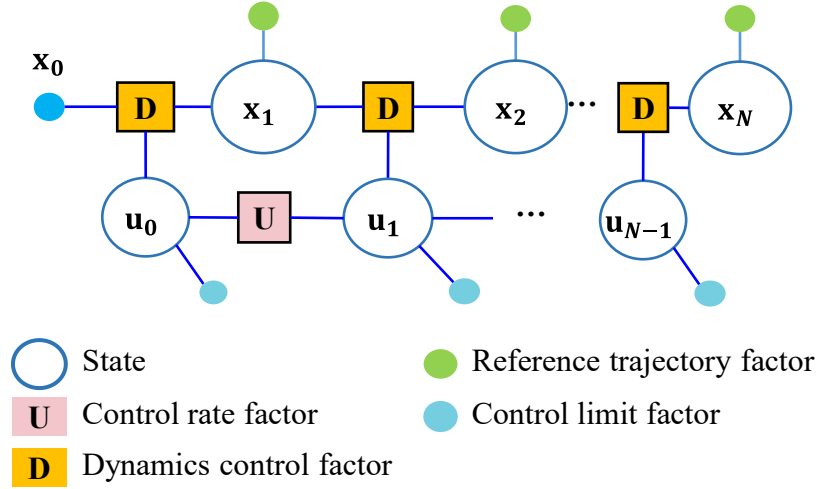

Fig. 3. Factor graph of MPC problem.

*1) Dynamics Control Factor*

The nominal dynamics model is updated according to Eq. (5), which considers the suction force when the quadrotor nears the wall. The dynamics model $\dot{\mathbf{x}} = \mathbf{f}(\mathbf{x}, \mathbf{u})$ of a quadrotor is improved as follows:

$$\begin{aligned} \dot{\mathbf{p}}_b^w &= \mathbf{v}_b^w \\ \dot{\mathbf{R}}_b^w &= \mathbf{R}_b^w \boldsymbol{\omega}_b^\times \\ \dot{\mathbf{v}}_b^w &= -\mathbf{e}_3 g + \frac{\mathbf{R}_b^w \left(\mathbf{e}_3 T_b + \mathbf{F}_{drag}^b\right) + \mathbf{F}_s}{m_b} \end{aligned} \quad (13)$$

The discrete state propagation function in the dynamic factor residuals is as follows:

$$\hat{\mathbf{x}}_{k+1} = \mathbf{F}(\mathbf{x}_k, \mathbf{u}_k) = \mathbf{x}_k \boxplus \Delta t f(\mathbf{x}_k, \mathbf{u}_k) \tag{14}$$

The dynamic factor residual $r^D$ is as follows:

$$r^D(\mathbf{x}_{k+1}, \mathbf{x}_k, \mathbf{u}_k) = \mathbf{x}_{k+1} \boxminus \hat{\mathbf{x}}_{k+1}$$

$$= \begin{bmatrix} \mathbf{p}^w_{b_{k+1}} - \mathbf{p}^w_{b_k} - \Delta t \mathbf{v}^w_{b_k} - 0.5 \dot{\mathbf{v}}^w_{b_k} \Delta t^2 \\ \text{Logmap}(\mathbf{R}^{b_k}_{b_{k+1}}) - \Delta t \boldsymbol{\omega}_{b_k} \\ \mathbf{v}^w_{b_{k+1}} - \mathbf{v}^w_{b_k} - \Delta t \dot{\mathbf{v}}^w_{b_k} \end{bmatrix} \tag{15}$$

$$\mathbf{R}^{b_k}_{b_{k+1}} = (\mathbf{R}^w_{b_k})^T \mathbf{R}^w_{b_{k+1}}$$

To propagate covariance, the dynamics model factor residuals can be represented in the body frame from the Eq. (15):

$$\bar{\mathbf{e}}_p = \mathbf{R}^{b_k}_w \mathbf{p}^k_p - 0.5 \frac{\mathbf{F}^{b_k}_{drag} + T_{b_k} \mathbf{e}_3}{m_b} \Delta t^2$$

$$\bar{\mathbf{e}}_\theta = \text{Log}(\mathbf{R}^{b_k}_w \mathbf{R}^w_{b_{k+1}}) - \boldsymbol{\omega}_{b_k} \Delta t \tag{16}$$

$$\bar{\mathbf{e}}_v = \mathbf{R}^{b_k}_w \mathbf{p}^k_v - \frac{D \mathbf{R}^b_w \mathbf{v}^w_b}{m_b} \Delta t - \frac{T_{b_k} \mathbf{e}_3}{m_b} \Delta t$$

where $\mathbf{p}^k_p$ and $\mathbf{p}^k_v$ are as follows:

$$\mathbf{p}^k_p = \mathbf{p}^w_{b_{k+1}} - \mathbf{v}^w_{b_k} \Delta t + 0.5 \mathbf{e}_3 g \Delta t^2 - 0.5 \frac{\mathbf{F}^k_s}{m_b} \Delta t^2 - \mathbf{p}^w_{b_k}$$

$$\mathbf{p}^k_v = \mathbf{v}^w_{b_{k+1}} - \mathbf{v}^w_{b_k} + \mathbf{e}_3 g \Delta t - \frac{\mathbf{F}^k_s}{m_b} \Delta t \tag{17}$$

The dynamics model uncertainty is assumed to mainly come from command's Gaussian noise. Therefore, the thrust and velocity control model are assumed to be,

$$T_{b_k} = \hat{T}_{b_k} - n^k_T, n^k_T \sim \mathcal{N}(0, \sigma_T)$$

$$\boldsymbol{\omega}_{b_k} = \hat{\boldsymbol{\omega}}_{b_k} - \mathbf{n}^k_\omega, \mathbf{n}^k_\omega \sim \mathcal{N}(0, \sigma_\omega \mathbf{I}_3) \tag{18}$$

where $\hat{T}_{b_k}$ and $\hat{\boldsymbol{\omega}}_{b_k}$ are control commands, $T_{b_k}$ and $\boldsymbol{\omega}_{b_k}$ are actual controlled thrust and angular velocity, $n_T$ and $\mathbf{n}_\omega$ are acceleration and angular velocity's Gaussian noise.

Based on Eq. (16), the dynamics factor residuals are specified as follows,

$$r^D(\mathbf{x}_{k+1}, \mathbf{x}_k, \mathbf{u}_k) = \mathbf{A}_k + \mathbf{B}(\mathbf{u}_k - \mathbf{w}_k),$$

$$\mathbf{A}_k = \begin{bmatrix} \mathbf{R}_w^{b_k}\mathbf{p}_p^k - 0.5\dfrac{\mathbf{D}\mathbf{R}_w^{b_k}\mathbf{v}_{b_k}^w}{m_b}\Delta t^2 \\ \mathrm{Log}\left(\mathbf{R}_w^{b_k}\mathbf{R}_{b_{k+1}}^w\right) \\ \mathbf{R}_w^{b_k}\mathbf{p}_v^k - \dfrac{\mathbf{D}\mathbf{R}_w^{b_k}\mathbf{v}_{b_k}^w}{m_b}\Delta t \end{bmatrix},$$

$$\mathbf{B} = -\begin{bmatrix} \dfrac{0.5\Delta t^2 \mathbf{I}_3}{m_b} & 0 \\ 0 & \Delta t \mathbf{I}_3 \\ \dfrac{\Delta t \mathbf{I}_3}{m_b} & 0 \end{bmatrix}, \quad \mathbf{u}_k = \begin{bmatrix} \hat{T}_{b_k} \\ \hat{\omega}_{b_k} \end{bmatrix}, \quad \mathbf{w}_k = \begin{bmatrix} \mathbf{n}_T^k \\ \mathbf{n}_\omega^k \end{bmatrix}$$

(19)

Furthermore, the covariance matrix $\mathbf{P}^D$ of $r^D$ is as follows:

$$\begin{aligned}\mathbf{P}^D &= \mathbf{B}\boldsymbol{\Sigma}_u \mathbf{B}^T \\ &= \begin{bmatrix} 0.25\Delta t^4 \sigma_a \mathbf{I}_3 / m_b^2 & 0 & 0.5\Delta t^3 \sigma_a \mathbf{I}_3 \\ 0 & \Delta t^2 \sigma_a \mathbf{I}_3 & 0 \\ 0.5\Delta t^3 \sigma_a \mathbf{I}_3 & 0 & \Delta t^2 \sigma_w \mathbf{I}_3 \end{bmatrix} \end{aligned}$$

(20)

The Jacobian matrix of $r^D$ concerning $\mathbf{x}_k$ is as follows:

$$\dfrac{\partial r^D}{\partial \mathbf{x}_k} = \begin{bmatrix} -\mathbf{R}_w^{b_k} & \left(\mathbf{R}_w^{b_k}\mathbf{p}_p^k\right)^\times & -\mathbf{R}_w^{b_k}\Delta t - 0.5\dfrac{\mathbf{D}\mathbf{R}_w^{b_k}}{m_b}\Delta t^2 \\ 0 & -\mathbf{R}_w^{b_{k+1}}\mathbf{R}_{b_k}^w & 0 \\ 0 & \dfrac{\partial \bar{\mathbf{e}}_v}{\partial \mathbf{R}_{b_k}^w} & \dfrac{\partial \bar{\mathbf{e}}_v}{\partial \mathbf{v}_{b_k}^w} \end{bmatrix}$$

(21)

where $\dfrac{\partial \bar{\mathbf{e}}_v}{\partial \mathbf{R}_{b_k}^w}$ and $\dfrac{\partial \bar{\mathbf{e}}_v}{\partial \mathbf{v}_{b_k}^w}$ are as follows:

$$\dfrac{\partial \bar{\mathbf{e}}_v}{\partial \mathbf{R}_{b_k}^w} = \left(\mathbf{R}_w^{b_k}\mathbf{p}_v^k\right)^\times - \mathbf{D}\left(\mathbf{R}_w^{b_k}\mathbf{v}_{b_k}^w\right)^\times \Delta t$$

$$\dfrac{\partial \bar{\mathbf{e}}_v}{\partial \mathbf{v}_{b_k}^w} = -\mathbf{R}_w^{b_k} - \dfrac{\mathbf{D}\mathbf{R}_w^{b_k}\Delta t}{m_b}$$

(22)

The Jacobian matrix of error $r^D$ to state $\mathbf{x}_{k+1}$ is as follows:

$$\dfrac{\partial r^D}{\partial \mathbf{x}_{k+1}} = \begin{bmatrix} \mathbf{R}_w^{b_k} & 0 & 0 \\ 0 & \mathbf{I}_3 & 0 \\ 0 & 0 & \mathbf{R}_w^{b_k} \end{bmatrix}$$

(23)

The Jacobian matrix of $r^D$ concerning $\mathbf{u}_k$ is as follows:

$$\frac{\partial \mathbf{r}^D}{\partial \mathbf{u}_k} = -\begin{bmatrix} \frac{0.5\Delta t^2 \mathbf{I}_3}{m_b} & 0 \\ 0 & \Delta t \mathbf{I}_3 \\ \frac{\Delta t \mathbf{I}_3}{m_b} & 0 \end{bmatrix} \tag{24}$$

*2) Reference Trajectory Factor*

The reference trajectory factor's residual is as follows:

$$\mathbf{r}^{ref} = \mathbf{x} \boxminus \mathbf{x}^{ref} = (\mathbf{p}, \mathbf{R}, \mathbf{v}) \boxminus (\mathbf{p}^{ref}, \mathbf{R}^{ref}, \mathbf{v}^{ref}) \tag{25}$$

where the position residual is $\mathbf{r}_p^{ref} = \mathbf{p} - \mathbf{p}^r$, the velocity residual is $\mathbf{r}_v^{ref} = \mathbf{v} - \mathbf{v}^r$, and the rotation residual is $\mathbf{r}_\theta^{ref} = \text{Logmap}(\mathbf{R}^T \mathbf{R}^r)$. Then, $\mathbf{r}^{ref} = \left[ (\mathbf{r}_p^{ref})^T \quad (\mathbf{r}_v^{ref})^T \quad (\mathbf{r}_\theta^{ref})^T \right]^T$.

*3) Control Rate Factor and Control Limit Factor*

The control boundary factor ensures that the control input meets the actuator's limits. Define a bounding function $\mathrm{h}(\mathbf{u}_k)$ for the control input inequality, which is as follows:

$$\mathrm{h}(\mathbf{u}_k) = \max(\mathbf{u}_k - \mathbf{u}_{max}, 0) + \max(\mathbf{u}_{min} - \mathbf{u}_k, 0) \tag{26}$$

where $\mathbf{u}_{min}$ is the input lower bound and $\mathbf{u}_{max}$ is the input upper bound.

To smooth the trajectory, the control rate factor is used to limit the rate of control inputs, which is as follows:

$$\mathbf{r}^U(\mathbf{u}_k, \mathbf{u}_{k+1}) = \mathbf{u}_k - \mathbf{u}_{k+1} \tag{27}$$

Finally, the Levenberg-Marquardt algorithm [9, 10] can be used to solve the MPC problem.

*3.4 Dynamic Identification*

While the suction force principle and model are derived through the force experiments, the aerodynamic configuration of a quadrotor in flight may differ significantly. Parameter identification is employed to accurately estimate the dynamics model parameters. $d_{thr}$ and $k_s$ require online identification. Based on Eq. (13) and Eq. (15), force residuals are defined as follows:

$$\mathbf{r}^F(\mathbf{F}_s) = m_b \mathbf{a}_b^w + m_b \mathbf{e}_3 g - \mathbf{R}_b^w (\mathbf{e}_3 T_b + \mathbf{F}_{drag}^b) - \mathbf{F}_s \tag{28}$$

The visual-inertial odometry (VIO) can estimate the acceleration $\mathbf{a}_b^w$, velocity $\mathbf{v}_b^w$, and rotation $\mathbf{R}_b^w$. Based on a period of observation and control input in the presence of suction, we construct numerous observation equations using Eq. (28). Then, parameters $d_{thr}$ and $k_s$ can be identified from flight data using Levenberg-Marquardt algorithm [9, 10].

The derivative of force residuals $\mathbf{r}^F$ to coefficient $k_s$ is as follows:

$$\frac{\partial \mathbf{r}^F}{\partial k_s} = -\mathbf{R}_{P_k}^w \sum_{j=0}^{3} \mathbf{e}_1 d_j \tag{29}$$

The derivative of $\mathbf{r}^F$ concerning $d_{thr}$ is as follows:

$$\frac{\partial \mathbf{r}^F}{\partial d_{thr}} = \begin{cases} 4\mathbf{R}_{P_k}^w \mathbf{e}_1 k_s & \mathbf{e}_1^T \mathbf{p}_{r_j}^{P_k} \geq 0 \\ -4\mathbf{R}_{P_k}^w \mathbf{e}_1 k_s & \mathbf{e}_1^T \mathbf{p}_{r_j}^{P_k} < 0 \end{cases} \qquad (30)$$

## 4. SIMULATIONS AND EXPERIMENTS

### 4.1 Simulation and Experiment Scheme

#### 4.1.1 Simulation Scheme

The quadrotor targets a circular trajectory along a wall, and control inputs to quadrotor added with Gaussian noise. The simulation framework is illustrated in Fig. 4. The thrust model is $T = T_{cmd} + n_T, n_T \sim \mathcal{N}(0, 0.2\text{N})$ and the angular velocity model is $\boldsymbol{\omega}_b = \boldsymbol{\omega}_b^{cmd} + \mathbf{n}_\omega, \mathbf{n}_\omega \sim \mathcal{N}(\mathbf{0}, \mathbf{I}_3 \cdot 0.2\text{rad/s})$. The wall-proximity disturbance is simulated based on the suction model. The proposed method is compared with cascaded PID and MPC to verify the feasibility. The suction force scale factor is set to 1.0 N/m, 4.0 N/m, or 10.0 N/m, respectively, and threshold distance is set to 10.0 cm.

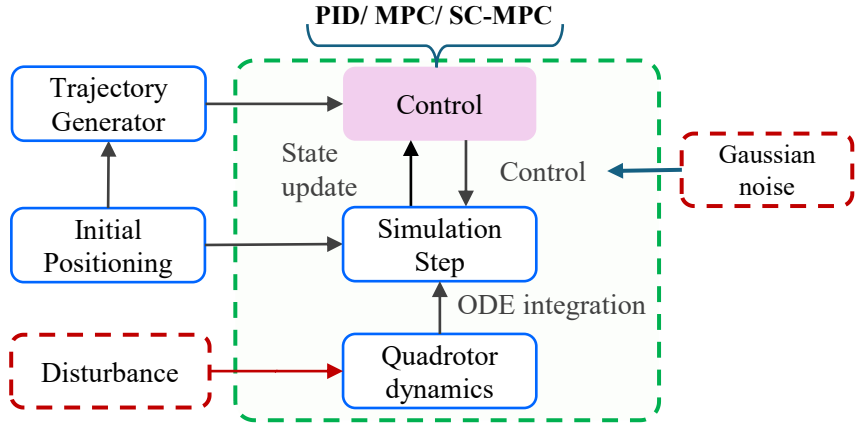

Fig. 4. Framework of the simulation system.

#### 4.1.2 Wall-proximity Effect Experiment Scheme

The suction force induced by the wall-proximity effect attracts the quadrotor toward the vertical surface, eventually colliding with the surface. A noticeable acceleration towards the vertical surface is observed. To validate this phenomenon and substantiate our hypothesis, a tabletop experiment has been designed, as depicted in Fig. 1-b. Two sets of experiments are conducted, to model the influence of wall distance and rotor speed on the suction forces exerted by the vertical surface. The wall-proximity effect is examined by incrementally moving a large vertical surface closer to the hovering quadrotor. Control input signals are programmed into the control board, which transmits them to the electronic speed controller (ESC) to regulate the rotor. Simultaneously, force data is collected using a data interpreter connected with a computer.

#### 4.1.3 Trajectory Following Experiment Scheme

Four ducts are installed to encase the rotors. A motion tracking system can provide real-time data on the quadrotor's pose and the wall's pose. The thrust estimation module computes the required throttle settings, while the low-level controller tracks the throttle and angular velocity commands. The performance of cascaded PID, MPC, and the proposed SC-MPC in trajectory tracking near the wall is evaluated and compared. The overall framework is illustrated in Fig. 5.

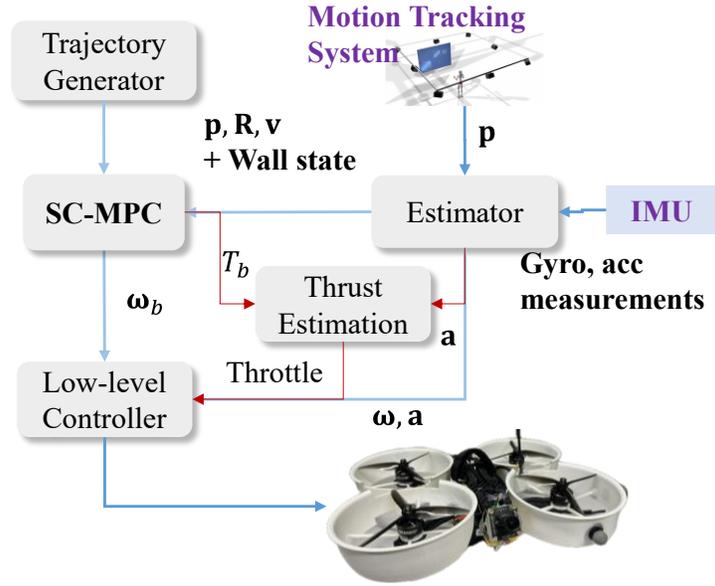

Fig. 5. Framework of the quadrotor system.

*4.2 Simulation Results*

As illustrated in Fig. 6, the trajectory tracking error curve of the quadrotor shows that the proposed method can alleviate the influence of wall-proximity effect, and the trajectory tracking accuracy is significantly better than cascaded MPC. For the bad case of k = 10.0 N/m, the proposed method (yellow curve) eliminates errors caused by near wall effects compared with MPC (purple curve). From the Table I, the larger the suction force scale factor, the greater the control error in the direction perpendicular to the wall.

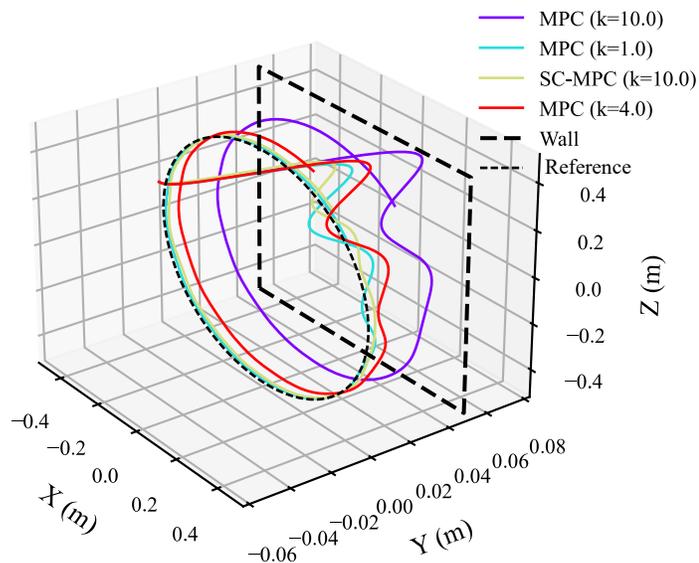

Fig. 6. MPC versus SC-MPC: Circular trajectory tracking position and tracking error curve for different parameters ($d_{thr} = 0.10\text{m}$).

Table I The RMSE of the X-axis and Y-axis position control for different parameters and methods.

| Method | $k_s$ (N/m) | $d_{thr}$ (cm) | MAE of Y (cm) | RMSE of Y (cm) |
|---|---|---|---|---|
| Cascaded PID | 10.0 | 10 cm | 0.041 | 0.054 |
| MPC | 1.0 | 10 | **0.0027** | 0.0047 |
| MPC | 4.0 | 10 | 0.0088 | 0.0098 |
| MPC | 10.0 | 10 | 0.029 | 0.030 |
| SC-MPC | 10.0 | 1 | 0.0032 | **0.0039** |

*4.3 Force Measurement and Modelling*

The duct height is 35 mm, and top opening diameter is 66 mm. As shown in Fig. 2, a three-blade rotor is used, secured onto a bracket connected to the force sensor. All ducts were manufactured from a PLA 3D printer. Their weigh approximately equal to 34g each. **Distance**: Experiments are also conducted at a constant PWM signal (PWM = 1350). Four distances are selected to measure the suction force. We quickly pushed the wall closer to the duct and measured its force at a specified distance. The magnitude of the suction force generated by the wall-proximity effect is obtained by calculating the change in force. **PWM**: Experiments are also conducted at four pulse width modulation (PWM) signals: 1155, 1255, 1355, and 1455. Most drones do not obtain rotors' rotational speed. In addition, for wall inspectors, its acceleration is generally very low, and it can be approximated that the rotational speed does not change much. Therefore, in our near-wall trajectory following experiments, PWM was ignored. In our previous research [15], the dynamics model of actuator is modelled that the rotor rotational speed $\omega \approx k_1 (PWM)^2 + b_1$.

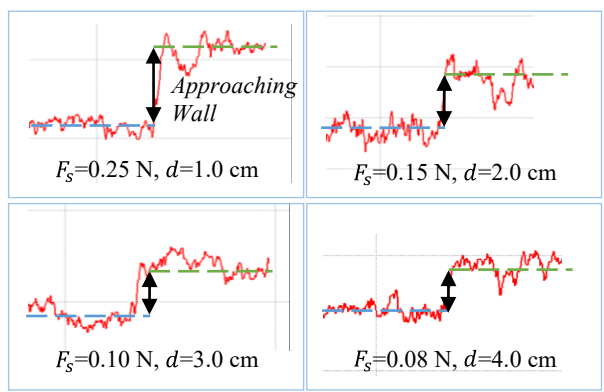

Fig. 7. Suction force measurements with different distances between the duct and the wall.

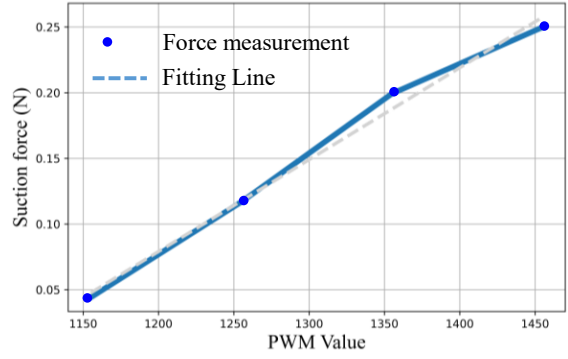

Fig. 8. Suction force measurements of varying PWM values: The blue line represents experimental results. The experiment is conducted at $d=1\text{cm}$ from the vertical surface with the duct installed.

*Conclusion of Suction Model*: (1) The suction force decreases linearly with distance $d$ as $F_s = k_s(d_{thr} - d)$ for $d < d_{thr}$, beyond which it vanishes ($F_s = 0$); (2) The suction force increases approximately linearly with the increase of PWM value.

*4.4 Parameters Identification and Trajectory Following in Wall-proximity*

The flight data is collected using cascaded PID control, which is used to solve suction parameters. The suction parameters of our quadrotor are $d_{thr} = 0.132\,\text{m}$, $k_s = 4.1\,\text{N/m}$. In theory, four ducted rotors can generate a maximum suction force of 1.08 N. Our quadrotor weighs 1 kg. The maximum suction force reaches 10.08% of gravity, indicating a substantial influence.

As illustrated in Fig. 9, the proposed method is compared with cascaded PID and MPC in terms of line tracking accuracy near a glass-made wall. The cascaded PID experiences the maximum amplitude of vibration, especially after the quadrotor hits the wall; its ability to maintain control stability is poor. MPC shows better robustness but still easily collides with walls. The proposed SC-MPC demonstrates extremely high precision wall following flight under wall-proximity effect and completely avoids collisions. This also indicates that the quadrotor can precisely fly along a wall at an extremely close distance.

In addition, the position tracking curves are shown in Fig. 10. The quadrotor maintains fixed X-axis and Y-axis positions and repeatedly moves from bottom to top. Collision not only causes oscillations in the X-axis and Y-axis positions but also deteriorates the accuracy of Z-axis position control. Therefore, collision is a sudden and significant disturbance that needs to be eliminated. For some real-world applications, such as building inspection and cleaning, the proposed method is a promising way to improve the safety of flight operations. The mean absolute error (MAE) and root mean squared errors (RMSE) are listed in Table II. The experimental results demonstrate the superior performance of our proposed SC-MPC method compared to both conventional cascaded PID control and standard MPC approaches. As shown in Table II, SC-MPC achieves significantly lower tracking errors in both X and Y axes. Specifically, for X-axis position control, SC-MPC reduces the RMSE by 74% (from 8.1 cm to 2.1 cm) compared to cascaded PID control and by 60% compared to standard MPC (from 5.3 cm to 2.1 cm). Similarly, for Y-axis control, SC-MPC shows a 79% improvement over cascaded PID (from 9.6 cm to 2.0 cm) and a 53% improvement over MPC (from 4.3 cm to 2.0 cm). The mean absolute errors (MAE) follow the same trend, with SC-MPC outperforming both baseline methods across all metrics. These quantitative results validate the effectiveness of our suction-compensated approach in achieving precise position control during near-wall flight operations.

In experiments, the proposed method achieves a control frequency of 100 Hz. The average time cost is 5.4 milliseconds, and 99.8% of the optimization problem solving is less than 10 milliseconds. The time cost of the traditional model is close to the proposed JPCM.

Table II    The RMSE of the X-axis and Y-axis position control.

| Method | MAE of X (cm) | RMSE of X (cm) | MAE of Y (cm) | RMSE of Y (cm) |
|---|---|---|---|---|
| cascaded PID | 4.5 | 8.1 | 6.2 | 9.6 |
| MPC | 1.7 | 5.3 | 2.1 | 4.3 |
| SC-MPC (ours) | 1.2 | 2.1 | 1.4 | 2.0 |

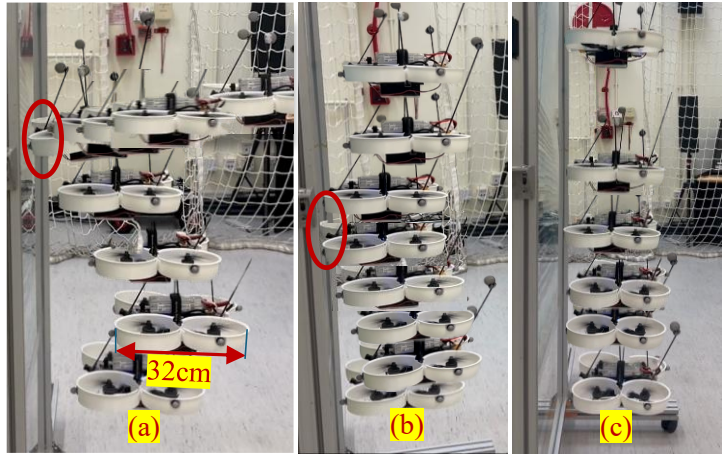

Fig. 9. Trajectory tracking comparison through video. (a) Cascaded PID; (b) MPC; (c) Proposed method SC-MPC. The red circle depicts the collision.

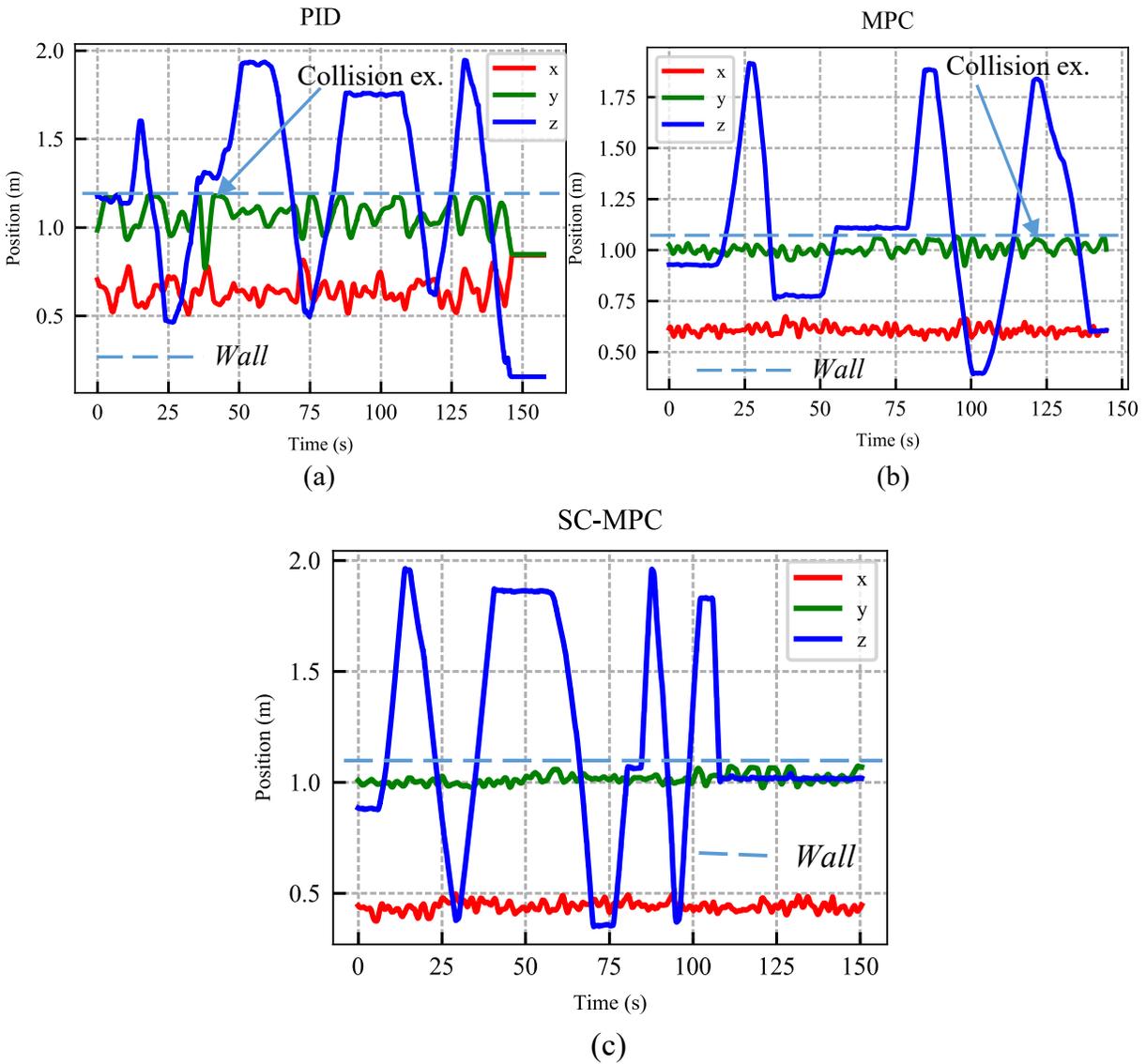

Fig. 10. Trajectory tracking comparison. (a) cascaded PID; (b) MPC; (c) Proposed method. Collision ex. represents collision example.

## 5. Conclusion and Future work

In many scenarios, such as urban or indoor environments, the lack of open space presents significant dangers for MAVs when they come close to walls. The wall-proximity effect generates a suction force on the MAV, which can cause it to shake violently or even collide with the wall. Thus, we model the wall-proximity effect using smog visualization and force measurements. Thereby, the dynamics model-compensated control method is presented to mitigate the wall-proximity effect. The simulation and trajectory tracking experiments demonstrate accurate and safe flights in wall-proximity.